\documentclass{article}

\PassOptionsToPackage{numbers, compress}{natbib}

\usepackage[final]{neurips_2024}

\usepackage[utf8]{inputenc} 
\usepackage[T1]{fontenc}    
\usepackage{hyperref}       
\usepackage{url}            
\usepackage{booktabs}       
\usepackage{amsfonts}       
\usepackage{nicefrac}       
\usepackage{microtype}      
\usepackage{xcolor}         
\usepackage{tikz}
\usepackage{graphicx}
\usepackage{amsmath}
\usepackage{algorithm}

\usepackage{listings}
\usepackage{bm}

\lstdefinestyle{overleaf}{
    backgroundcolor=\color[rgb]{0.95,0.95,0.92},   
    commentstyle=\color[rgb]{0,0.6,0},
    keywordstyle=\color{magenta},
    numberstyle=\tiny\color[rgb]{0.5,0.5,0.5},
    stringstyle=\color[rgb]{0.58,0,0.82},
    basicstyle=\ttfamily\footnotesize,
    breakatwhitespace=false,         
    breaklines=true,                 
    captionpos=b,                    
    keepspaces=true,                 
    numbers=left,                    
    numbersep=5pt,                  
    showspaces=false,                
    showstringspaces=false,
    showtabs=false,                  
    tabsize=2
}

\lstdefinestyle{simple}{
  backgroundcolor=\color{white},
  basicstyle=\fontsize{7.5pt}{7.5pt}\ttfamily\selectfont,
  keywordstyle=\fontsize{7.5pt}{7.5pt}\ttfamily\selectfont,
}
\author{%
  Ömer Veysel Çağatan \\
  Department of Computer Engineering\\
  Koç University \\
  Sarıyer, İstanbul 34450 \\
  \texttt{ocagatan19@ku.edu.tr} \\
}

\begin{document}

\title{SigCLR: Sigmoid Contrastive Learning of Visual Representations}
\maketitle

\begin{abstract}
We propose SigCLR: Sigmoid Contrastive Learning of Visual Representations. SigCLR utilizes the logistic loss that only operates on pairs and does not require a global view as in the cross-entropy loss used in SimCLR. We show that logistic loss shows competitive performance on CIFAR-10, CIFAR-100, and Tiny-IN compared to other established SSL objectives. Our findings verify the importance of learnable bias as in the case of SigLUP, however, it requires a fixed temperature as in the SimCLR to excel. Overall, SigCLR is a promising replacement for the SimCLR which is ubiquitous and has shown tremendous success in various domains. 
  
\end{abstract}

\section{Introduction}

Contrastive learning has been widely adopted as a pretraining objective in vision tasks to develop universal vision backbones, addressing the scalability challenges of supervised learning due to the high labeling costs ~\cite{chen2020simple, chen2020big, dwibedi2021little, he2020momentum, li2021prototypical, mitrovic2020representation, yeh2022decoupled}. This approach not only eliminates the need for labeled data but also allows models to be trained on large-scale image datasets ~\cite{oquab2023dinov2}, aiming to replicate the success seen in the language domain ~\cite{touvron2023llama, openai2023gpt4}. However, a significant limitation of contrastive learning is its reliance on negative samples, which necessitates large batch sizes ~\cite{chen2020simple}. In contrast, non-contrastive methods ~\cite{bardes2022vicreg, caron2021unsupervised, chen2020exploring, ermolov2021whitening, grill2020bootstrap, ozsoy2022selfsupervised, zbontar2021barlow, caron2021emerging} focus solely on positive pairs and tend to perform well, particularly in scenarios where smaller batch sizes are used.

Although non-contrastive objectives have a clear advantage over contrastive objectives in vision pretraining, their usage is limited~\cite{schwarzer2023biggerbetterfasterhumanlevel,zhou2022noncontrastivelearningmeetslanguageimage,çağatan2024unseeunsupervisednoncontrastivesentence}. On the other hand contrastive learning has been extensively used and shown to excel as a domain-agnostic representation learning objective. This includes language~\cite{fang2020cert,gao2022simcse,su2022contrastive,pan2021contrastive,meng2021cocolm}, language-image~\cite{radford2021learning,zhai2023sigmoid,li2022blip}, graphs~\cite{you2021graph}, biology~\cite{Wang_2022}, chemistry~\cite{Sanchez-Fernandez2022.11.17.516915} and medicine~\cite{wang2022medclip,chaitanya2020contrastive}.

SSL models that are cast under contrastive learning usually modify InfoNCE~ objective\cite{oord2019representation} which can also be described as softmax contrastive loss. A simple alternative to softmax contrastive loss is to use sigmoid loss and treat every pair as a binary classification problem. Even though it has a very straightforward recipe, it has been shown to be inferior to softmax objective in image pretraining~\cite{chen2020simple,hjelm2019learning}. SigLIP~\citet{zhai2023sigmoid} revisits the sigmoid objective and demonstrates that it is a more efficient and equally performant model as the CLIP~\cite{radford2021learning} which employs the softmax contrastive objective. The central innovation in the SigLIP~\cite{zhai2023sigmoid} lies in its adoption of learnable temperature and bias. This stands in contrast to the softmax approach, which relies solely on a fixed temperature in its loss formulation.

An important consideration is whether the success of SigLIP can be applied to vision pretraining. With this in mind, we revisit the sigmoid loss in the context of vision pretraining, where it falls short compared to softmax. To explore this further, we introduce SigCLR: Sigmoid Contrastive Learning of Visual Representations to investigate the potential of sigmoid-based contrastive pretraining for vision.

Our findings are consistent with SigLIP which emphasizes the significance of learnable bias for the sigmoid loss. During initialization, the substantial imbalance introduced by a multitude of negative samples prevails in the loss, prompting the need for substantial initial optimization steps to address this inherent bias~\cite{zhai2023sigmoid}. Likewise, We observe the same phenomenon in vision pretraining and therefore utilize a learnable bias. The addition of learnable bias significantly enhances the performance of sigmoid loss, however, a fixed temperature is necessary to achieve the pinnacle performance of SigCLR. 

We test SigCLR in widely used benchmarks such as CIFAR-10, and CIFAR100~\cite{Krizhevsky2009LearningML}, Tiny ImageNet~\cite{Le2015TinyIV} and ImageNet 100~\cite{Le2015TinyIV} always outperforms SimCLR and have competitive results with other objectives. Thus, sigmoid contrastive learning stands as a simple yet powerful objective that later can be extended to other domains in which softmax is used. 
\begin{figure}[!t]
\small
    \centering
\begin{tikzpicture}
    \node at (0,1.8) (h) {$\longleftarrow\,$Encoder Embeddings$\,\longrightarrow$};
    \node[draw, circle] at (0,-1) (x) {$\,~\bm{x}~\,$};
    \node[draw, circle] at (-2.5,0) (x1) {$\tilde{\bm{x}}_i$};
    \node[draw, circle] at (2.5,0) (x2) {$\tilde{\bm{x}}_j$};
    \node at (-2.5,1.8) (h) {$\bm h_i$};
    \node at (2.5,1.8) (c) {$\bm h_j$};
    \node at (-2.5,3) (hh) {$\bm z_i$};
    \node at (2.5,3) (cc) {$\bm z_j$};
    \path[->] 
        (x)  edge [>=latex] node[below,rotate=-25] {$t\sim\mathcal{T}$} (x1)
        (x)  edge [>=latex] node[below,rotate=25] {$t'\sim \mathcal{T}$} (x2)
        (x1)  edge [>=latex] node[left,rotate=0] {$f(\cdot)$} (h)
        (x2)  edge [>=latex] node[right,rotate=0] {$f(\cdot)$} (c)
        (h)  edge [>=latex] node[left,rotate=0] {$g(\cdot)$} (hh)
        (c)  edge [>=latex] node[right,rotate=0] {$g(\cdot)$} (cc);
    \path[<->]
        (hh)  edge [>=latex] node[above,rotate=0] {Sigmoid Contrastive Loss} (cc);
    \end{tikzpicture}
    \caption{SigCLR follows a highly similar setup as SimCLR~\cite{chen2020simple}. We randomly select two distinct data augmentation operators, denoted as $t\sim \mathcal{T}$ and $t'\sim \mathcal{T}$, from the same family of augmentations from ~\cite{ozsoy2022selfsupervised}. These operators are applied independently to each data example, resulting in two correlated views. The training process involves optimizing a base encoder network, denoted as $f(\cdot)$, and a projection head, denoted as $g(\cdot)$, to maximize agreement between the representations produced by the augmented views. This optimization is achieved through the utilization of a sigmoid contrastive loss. Upon completion of training, the projection head is discarded and only encoder embeddings are utilized for evaluations.}
    \label{fig:framework}
\end{figure}
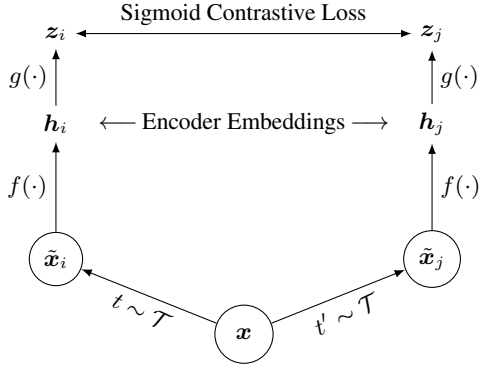

\section{Method}\label{sect:method}

SigCLR learns representations by maximizing the similarity between distinctively augmented perspectives of a given data instance by applying the sigmoid contrastive loss within the latent space. We present the illustration of SigCLR in Figure \ref{fig:framework}.  This alternative eliminates the need for calculating global normalization factors. The sigmoid-based loss operates independently on each image pair, transforming the learning task into a standard binary classification across all pair combinations in the dataset. Positive labels are assigned to pairs consisting of augmented views, while negative labels are assigned to all other pairs. The formulation of the loss is as follows:
$$
-\frac 1 {|2\mathcal{B}|} \sum_{i=1}^{|2\mathcal{B}|} \sum_{j=1}^{|2\mathcal{B}|} \underbrace{ {k_{ij}}  \log\frac 1 {1+e^{z_{ij}(-t\mathbf{x}_i \cdot \mathbf{y}_j  - b)}}}_{\mathcal{L}_{ij}}
$$
where $z_{ij}$ is the label for a given image inputs, which equals 1 if they are positive, and $-1$ otherwise. $k_{ij}$ is 0 when $x_{i}$ and $y_{j}$ are the same view otherwise 1. Furthermore, We provide pseudocode in Algorithm \ref{alg:sigclr}.

\begin{table}[t]
  \caption{Top-1 accuracies {($\%$)} under linear evaluation on different datasets. Results are reported from \citep{JMLR:v23:21-1155,ozsoy2022selfsupervised}. We \textbf{bold} all top results that are statistically indistinguishable.} 
  \label{evaluation-table-1}
  \centering
  \small
  \footnotesize
  \begin{tabular}{@{}l@{\hspace*{1mm}}ccccc@{\hspace*{2mm}}ccc@{}}
    \toprule
    {Method} & \multicolumn{1}{c}{CIFAR-10} &  \multicolumn{1}{c}{CIFAR-100} & \multicolumn{1}{c}{Tiny-IN} \\
    \cmidrule(l){2-2} \cmidrule(l){3-3} \cmidrule(l){4-5}  \cmidrule(l){6-7} \cmidrule(l){8-8}
    & {ResNet-18} & {ResNet-18} & {ResNet-18}  &  &\\
    \midrule 
    SimCLR \citep{chen2020simple} & 90.74 & 65.78 & 48.84  \\
    SimSiam \citep{chen2020exploring}& 91.40 & 66.04& -  \\
    BYOL \citep{grill2020bootstrap} & 92.58 &  70.46 & 51.00  &\\
    W-MSE 2 \citep{ermolov2021whitening} & 91.55 & 66.10 & 49.22 &\\
    MoCo-V2 \citep{chen2020improved} & \textbf{92.94} & 69.89 &  - &  \\
    Barlow \citep{zbontar2021barlow} & 92.10 & 70.90 & - &  \\
    VICReg \citep{bardes2022vicreg} & 92.07 & 68.54& - & \\
    \midrule
    SimCLR* & 91.69 & 65.49 & 48.16  \\
    SigCLR & 91.77 & 66.98 & 48.94           \\
    
    \bottomrule
  \end{tabular}
\end{table}

\section{Experimental Setup}
We perform our experiments on CIFAR-10, CIFAR-100~\cite{Krizhevsky2009LearningML}, Tiny-IN~\cite{Le2015TinyIV}.

The experimental process consists of two successive phases: pretraining and linear evaluation. Initially, we employ the unsupervised pretraining of the encoder network using the proposed SigCLR method outlined in Section \ref{sect:method} on the training dataset. Once pretraining is completed, we proceed with linear evaluation—a standardized protocol used to assess the quality of the acquired representations, as established in prior research~\cite{chen2020simple,grill2020bootstrap}.

We employ a variant of the ResNet-18~\cite{he2015deep}, similar to the configurations described in SimCLR~\cite{chen2020simple}. The projector network is a three-layer MLP with ReLU activation functions for the hidden layers and linear activation functions for the output layer. The dimensions of the CIFAR-10, CIFAR-100, and Tiny-IN datasets are 1024-1024-128.

In the pretraining phase, we create two augmented versions of each input image by applying random cropping, resizing, and diverse operations including horizontal mirroring, color jittering, grayscale conversion, and Gaussian blurring. The augmentation parameters follow those of ~\cite{ozsoy2022selfsupervised,grill2020bootstrap}. In the training phase of linear evaluation, a single augmentation comprises random cropping, resizing, and a horizontal flip. During the test phase of linear evaluation, we use resizing and center crop augmentations, similar to the approach in \cite{bardes2022vicreg,zbontar2021barlow}.

We employ 1000 epochs of pretraining on CIFAR-10, CIFAR-100, Tiny-IN. We use Lars~\cite{you2017large} with cosine annealing and linear warmup, starting with a learning rate of 0.3 and linearly scaling with a starting batch size of 64. We incorporate a 10-epoch linear warmup with cosine annealing~\cite{loshchilov2017sgdrstochasticgradientdescent} and apply a weight decay of 0.000001. 

Since we employ different data augmentations than the original SimCLR, we also reproduced SimCLR on CIFAR-10, CIFAR-100, and Tiny-IN. We report these results as SimCLR* in table ~\ref{evaluation-table-1} and also ablations in table ~\ref{tab:model-simclr-results}.

\begin{table}[h]
\caption{\textbf{Effect of bias and temperature}~(CIFAR10 linear evaluation accuracy). In case of learnable bias, we initialize it to be -10. In all experiments batch size is 128.}
\label{tab:temperature_comparison}
\centering
\begin{tabular}{lcccc}
\toprule
\textbf{} & \textbf{Temp. = 1} & \textbf{Temp. = 2} & \textbf{Temp. = 5} & \textbf{Temp. = 10} \\
\midrule
Fixed Temp. w/ Learnable Bias & 90.76&91.25 & 91.53& 89.80 \\
Learnable Temp. w/ Learnable Bias & 84.16 &84.11 &84.07 & 84.67 \\
Fixed Temp. No Bias & 87.17&84.22 &27.15 &17.37 \\
\bottomrule
\end{tabular}
\end{table}
\newpage
\section{Results}

\citet{chen2020simple} demonstrated that temperature-normalized logistic loss significantly underperforms compared to temperature-normalized softmax loss on ImageNet. To further investigate, we reevaluated this setup on CIFAR-10 and observed that models with lower temperatures still lag behind established SSL objectives, while higher temperatures lead to notably poor performance.

Additionally, we examined the most effective approach from SigLIP~\cite{zhai2023sigmoid}, which recommends both learnable temperature and bias. While the learnable bias helps maintain reasonable performance across different temperatures, the best performance achieved remains inferior to that of the standard logistic loss.

Consequently, we implement a setup using a learnable bias to maintain consistent performance across temperatures, paired with a fixed temperature to optimize results. With these two straightforward modifications, we find that the logistic loss achieves performance comparable to state-of-the-art SSL objectives.

We further extend the evaluation of SigCLR to CIFAR-100 and Tiny-IN, where it shows competitive performance, highlighting the effectiveness of the sigmoid contrastive loss.

Lastly, we assess both SimCLR* and SigCLR across varying batch sizes to investigate whether the large batch size issues seen in softmax-based contrastive learning are present. Our results indicate that SigCLR performs exceptionally well with smaller batch sizes, aligning with the findings of SigLIP.

\begin{table}[t]
\centering
\caption{Results of SimCLR* across Datasets and Batch Sizes}
\label{tab:model-simclr-results}
\begin{tabular}{@{}llllll@{}}
\toprule
& \multicolumn{5}{c}{Batch Size} \\
\cmidrule(l){2-6}
Dataset & {64} & {128} & {256} & {512} & {1024} \\
\midrule
CIFAR-10  & {90.56} & {91.69} & {92.23} & {92.42} & {92.26} \\
CIFAR-100  & {62.85} & {65.49} & {66.67} & {67.26} & {66.49} \\
Tiny-IN  & {46.08} & {48.16} & {49.92} & {49.16} & {49.94} \\
\bottomrule
\end{tabular}
\end{table}

\begin{table}[t]
\centering
\caption{Results of SigCLR across Datasets and Batch Sizes}
\label{tab:model-sigclr-results}
\begin{tabular}{@{}llllll@{}}
\toprule
& \multicolumn{5}{c}{Batch Size} \\
\cmidrule(l){2-6}
Dataset & {64} & {128} & {256} & {512} & {1024} \\
\midrule
CIFAR-10  & {91.26} & {91.77} & {92.11} & {92.59} & {92.62} \\
CIFAR-100  & {66.52} & {66.98} & {67.86} & {68.57} & {68.58} \\
Tiny-IN  & {47.53} & {48.94} & {49.62} & {50.56} & {51.54} \\
\bottomrule
\end{tabular}
\end{table}
\section{Future Work and Limitations}
Achieving results on ImageNet is crucial for establishing the credibility of SSL objectives. Therefore, the lack of such results considerably reduces the impact of our work. A key focus of future work is to extend our experiments to the final dataset, ImageNet-1k, as we believe this will provide a stronger foundation for our findings. However, the main limitation remains the significant computational resources required to train and evaluate this dataset.

\section{Conclusion}
We introduce SigCLR, a performant contrastive vision pretraining objective that employs sigmoid instead of the conventional softmax utilized in SimCLR. SigCLR demonstrates notable performance on established benchmarks, establishing itself as a strong contrastive objective. The straightforward objective of SigCLR eliminates the need for global normalization, a bottleneck for softmax, resulting in a highly efficient contrastive loss with enhanced performance.

\bibliography{Styles/neurips_2024}

\begin{thebibliography}{47}
\expandafter\ifx\csname natexlab\endcsname\relax\def\natexlab#1{#1}\fi

\bibitem[{Bardes et~al.(2022)Bardes, Ponce, and LeCun}]{bardes2022vicreg}
Adrien Bardes, Jean Ponce, and Yann LeCun. 2022.
\newblock \href {http://arxiv.org/abs/2105.04906} {VICReg: Variance-Invariance-Covariance Regularization for Self-Supervised Learning}.

\bibitem[{Beyer et~al.(2020)Beyer, Hénaff, Kolesnikov, Zhai, and van~den Oord}]{beyer2020imagenet}
Lucas Beyer, Olivier~J. Hénaff, Alexander Kolesnikov, Xiaohua Zhai, and Aäron van~den Oord. 2020.
\newblock \href {http://arxiv.org/abs/2006.07159} {Are we done with ImageNet?}

\bibitem[{Caron et~al.(2019)Caron, Bojanowski, Joulin, and Douze}]{caron2019deep}
Mathilde Caron, Piotr Bojanowski, Armand Joulin, and Matthijs Douze. 2019.
\newblock \href {http://arxiv.org/abs/1807.05520} {Deep Clustering for Unsupervised Learning of Visual Features}.

\bibitem[{Caron et~al.(2021{\natexlab{a}})Caron, Misra, Mairal, Goyal, Bojanowski, and Joulin}]{caron2021unsupervised}
Mathilde Caron, Ishan Misra, Julien Mairal, Priya Goyal, Piotr Bojanowski, and Armand Joulin. 2021{\natexlab{a}}.
\newblock \href {http://arxiv.org/abs/2006.09882} {Unsupervised Learning of Visual Features by Contrasting Cluster Assignments}.

\bibitem[{Caron et~al.(2021{\natexlab{b}})Caron, Touvron, Misra, Jégou, Mairal, Bojanowski, and Joulin}]{caron2021emerging}
Mathilde Caron, Hugo Touvron, Ishan Misra, Hervé Jégou, Julien Mairal, Piotr Bojanowski, and Armand Joulin. 2021{\natexlab{b}}.
\newblock \href {http://arxiv.org/abs/2104.14294} {Emerging Properties in Self-Supervised Vision Transformers}.

\bibitem[{Chaitanya et~al.(2020)Chaitanya, Erdil, Karani, and Konukoglu}]{chaitanya2020contrastive}
Krishna Chaitanya, Ertunc Erdil, Neerav Karani, and Ender Konukoglu. 2020.
\newblock \href {http://arxiv.org/abs/2006.10511} {Contrastive learning of global and local features for medical image segmentation with limited annotations}.

\bibitem[{Chen et~al.(2020{\natexlab{a}})Chen, Kornblith, Norouzi, and Hinton}]{chen2020simple}
Ting Chen, Simon Kornblith, Mohammad Norouzi, and Geoffrey Hinton. 2020{\natexlab{a}}.
\newblock \href {http://arxiv.org/abs/2002.05709} {A Simple Framework for Contrastive Learning of Visual Representations}.

\bibitem[{Chen et~al.(2020{\natexlab{b}})Chen, Kornblith, Swersky, Norouzi, and Hinton}]{chen2020big}
Ting Chen, Simon Kornblith, Kevin Swersky, Mohammad Norouzi, and Geoffrey Hinton. 2020{\natexlab{b}}.
\newblock Big Self-Supervised Models are Strong Semi-Supervised Learners.
\newblock \emph{arXiv preprint arXiv:2006.10029}.

\bibitem[{Chen et~al.(2020{\natexlab{c}})Chen, Fan, Girshick, and He}]{chen2020improved}
Xinlei Chen, Haoqi Fan, Ross Girshick, and Kaiming He. 2020{\natexlab{c}}.
\newblock \href {http://arxiv.org/abs/2003.04297} {Improved Baselines with Momentum Contrastive Learning}.

\bibitem[{Chen and He(2020)}]{chen2020exploring}
Xinlei Chen and Kaiming He. 2020.
\newblock \href {http://arxiv.org/abs/2011.10566} {Exploring Simple Siamese Representation Learning}.

\bibitem[{da~Costa et~al.(2022)da~Costa, Fini, Nabi, Sebe, and Ricci}]{JMLR:v23:21-1155}
Victor Guilherme~Turrisi da~Costa, Enrico Fini, Moin Nabi, Nicu Sebe, and Elisa Ricci. 2022.
\newblock \href {http://jmlr.org/papers/v23/21-1155.html} {solo-learn: A Library of Self-supervised Methods for Visual Representation Learning}.
\newblock \emph{Journal of Machine Learning Research}, 23(56):1--6.

\bibitem[{Dwibedi et~al.(2021)Dwibedi, Aytar, Tompson, Sermanet, and Zisserman}]{dwibedi2021little}
Debidatta Dwibedi, Yusuf Aytar, Jonathan Tompson, Pierre Sermanet, and Andrew Zisserman. 2021.
\newblock \href {http://arxiv.org/abs/2104.14548} {With a Little Help from My Friends: Nearest-Neighbor Contrastive Learning of Visual Representations}.

\bibitem[{Ermolov et~al.(2021)Ermolov, Siarohin, Sangineto, and Sebe}]{ermolov2021whitening}
Aleksandr Ermolov, Aliaksandr Siarohin, Enver Sangineto, and Nicu Sebe. 2021.
\newblock \href {http://arxiv.org/abs/2007.06346} {Whitening for Self-Supervised Representation Learning}.

\bibitem[{Fang et~al.(2020)Fang, Wang, Zhou, Ding, and Xie}]{fang2020cert}
Hongchao Fang, Sicheng Wang, Meng Zhou, Jiayuan Ding, and Pengtao Xie. 2020.
\newblock \href {http://arxiv.org/abs/2005.12766} {CERT: Contrastive Self-supervised Learning for Language Understanding}.

\bibitem[{Gao et~al.(2022)Gao, Yao, and Chen}]{gao2022simcse}
Tianyu Gao, Xingcheng Yao, and Danqi Chen. 2022.
\newblock \href {http://arxiv.org/abs/2104.08821} {SimCSE: Simple Contrastive Learning of Sentence Embeddings}.

\bibitem[{Grill et~al.(2020)Grill, Strub, Altché, Tallec, Richemond, Buchatskaya, Doersch, Pires, Guo, Azar, Piot, Kavukcuoglu, Munos, and Valko}]{grill2020bootstrap}
Jean-Bastien Grill, Florian Strub, Florent Altché, Corentin Tallec, Pierre~H. Richemond, Elena Buchatskaya, Carl Doersch, Bernardo~Avila Pires, Zhaohan~Daniel Guo, Mohammad~Gheshlaghi Azar, Bilal Piot, Koray Kavukcuoglu, Rémi Munos, and Michal Valko. 2020.
\newblock \href {http://arxiv.org/abs/2006.07733} {Bootstrap your own latent: A new approach to self-supervised Learning}.

\bibitem[{Hadsell et~al.(2006)Hadsell, Chopra, and LeCun}]{Hadsell2006DimensionalityRB}
Raia Hadsell, Sumit Chopra, and Yann LeCun. 2006.
\newblock \href {https://api.semanticscholar.org/CorpusID:8281592} {Dimensionality Reduction by Learning an Invariant Mapping}.
\newblock \emph{2006 IEEE Computer Society Conference on Computer Vision and Pattern Recognition (CVPR'06)}, 2:1735--1742.

\bibitem[{He et~al.(2020)He, Fan, Wu, Xie, and Girshick}]{he2020momentum}
Kaiming He, Haoqi Fan, Yuxin Wu, Saining Xie, and Ross Girshick. 2020.
\newblock \href {http://arxiv.org/abs/1911.05722} {Momentum Contrast for Unsupervised Visual Representation Learning}.

\bibitem[{He et~al.(2015)He, Zhang, Ren, and Sun}]{he2015deep}
Kaiming He, Xiangyu Zhang, Shaoqing Ren, and Jian Sun. 2015.
\newblock \href {http://arxiv.org/abs/1512.03385} {Deep Residual Learning for Image Recognition}.

\bibitem[{Hjelm et~al.(2019)Hjelm, Fedorov, Lavoie-Marchildon, Grewal, Bachman, Trischler, and Bengio}]{hjelm2019learning}
R~Devon Hjelm, Alex Fedorov, Samuel Lavoie-Marchildon, Karan Grewal, Phil Bachman, Adam Trischler, and Yoshua Bengio. 2019.
\newblock \href {http://arxiv.org/abs/1808.06670} {Learning deep representations by mutual information estimation and maximization}.

\bibitem[{Krizhevsky(2009)}]{Krizhevsky2009LearningML}
Alex Krizhevsky. 2009.
\newblock \href {https://api.semanticscholar.org/CorpusID:18268744} {Learning Multiple Layers of Features from Tiny Images}.

\bibitem[{Le and Yang(2015)}]{Le2015TinyIV}
Ya~Le and Xuan~S. Yang. 2015.
\newblock \href {https://api.semanticscholar.org/CorpusID:16664790} {Tiny ImageNet Visual Recognition Challenge}.

\bibitem[{Li et~al.(2022)Li, Li, Xiong, and Hoi}]{li2022blip}
Junnan Li, Dongxu Li, Caiming Xiong, and Steven Hoi. 2022.
\newblock \href {http://arxiv.org/abs/2201.12086} {BLIP: Bootstrapping Language-Image Pre-training for Unified Vision-Language Understanding and Generation}.

\bibitem[{Li et~al.(2021)Li, Zhou, Xiong, and Hoi}]{li2021prototypical}
Junnan Li, Pan Zhou, Caiming Xiong, and Steven C.~H. Hoi. 2021.
\newblock \href {http://arxiv.org/abs/2005.04966} {Prototypical Contrastive Learning of Unsupervised Representations}.

\bibitem[{Loshchilov and Hutter(2017)}]{loshchilov2017sgdrstochasticgradientdescent}
Ilya Loshchilov and Frank Hutter. 2017.
\newblock \href {http://arxiv.org/abs/1608.03983} {SGDR: Stochastic Gradient Descent with Warm Restarts}.

\bibitem[{Meng et~al.(2021)Meng, Xiong, Bajaj, Tiwary, Bennett, Han, and Song}]{meng2021cocolm}
Yu~Meng, Chenyan Xiong, Payal Bajaj, Saurabh Tiwary, Paul Bennett, Jiawei Han, and Xia Song. 2021.
\newblock \href {http://arxiv.org/abs/2102.08473} {COCO-LM: Correcting and Contrasting Text Sequences for Language Model Pretraining}.

\bibitem[{Mitrovic et~al.(2020)Mitrovic, McWilliams, Walker, Buesing, and Blundell}]{mitrovic2020representation}
Jovana Mitrovic, Brian McWilliams, Jacob Walker, Lars Buesing, and Charles Blundell. 2020.
\newblock \href {http://arxiv.org/abs/2010.07922} {Representation Learning via Invariant Causal Mechanisms}.

\bibitem[{OpenAI(2023)}]{openai2023gpt4}
OpenAI. 2023.
\newblock \href {http://arxiv.org/abs/2303.08774} {GPT-4 Technical Report}.

\bibitem[{Oquab et~al.(2023)Oquab, Darcet, Moutakanni, Vo, Szafraniec, Khalidov, Fernandez, Haziza, Massa, El-Nouby, Assran, Ballas, Galuba, Howes, Huang, Li, Misra, Rabbat, Sharma, Synnaeve, Xu, Jegou, Mairal, Labatut, Joulin, and Bojanowski}]{oquab2023dinov2}
Maxime Oquab, Timothée Darcet, Théo Moutakanni, Huy Vo, Marc Szafraniec, Vasil Khalidov, Pierre Fernandez, Daniel Haziza, Francisco Massa, Alaaeldin El-Nouby, Mahmoud Assran, Nicolas Ballas, Wojciech Galuba, Russell Howes, Po-Yao Huang, Shang-Wen Li, Ishan Misra, Michael Rabbat, Vasu Sharma, Gabriel Synnaeve, Hu~Xu, Hervé Jegou, Julien Mairal, Patrick Labatut, Armand Joulin, and Piotr Bojanowski. 2023.
\newblock \href {http://arxiv.org/abs/2304.07193} {DINOv2: Learning Robust Visual Features without Supervision}.

\bibitem[{Ozsoy et~al.(2022)Ozsoy, Hamdan, Arik, Yuret, and Erdogan}]{ozsoy2022selfsupervised}
Serdar Ozsoy, Shadi Hamdan, Sercan~Ö. Arik, Deniz Yuret, and Alper~T. Erdogan. 2022.
\newblock \href {http://arxiv.org/abs/2209.07999} {Self-Supervised Learning with an Information Maximization Criterion}.

\bibitem[{Pan et~al.(2021)Pan, Wang, Wu, and Li}]{pan2021contrastive}
Xiao Pan, Mingxuan Wang, Liwei Wu, and Lei Li. 2021.
\newblock \href {http://arxiv.org/abs/2105.09501} {Contrastive Learning for Many-to-many Multilingual Neural Machine Translation}.

\bibitem[{Radford et~al.(2021)Radford, Kim, Hallacy, Ramesh, Goh, Agarwal, Sastry, Askell, Mishkin, Clark, Krueger, and Sutskever}]{radford2021learning}
Alec Radford, Jong~Wook Kim, Chris Hallacy, Aditya Ramesh, Gabriel Goh, Sandhini Agarwal, Girish Sastry, Amanda Askell, Pamela Mishkin, Jack Clark, Gretchen Krueger, and Ilya Sutskever. 2021.
\newblock \href {http://arxiv.org/abs/2103.00020} {Learning Transferable Visual Models From Natural Language Supervision}.

\bibitem[{Sanchez-Fernandez et~al.(2023)Sanchez-Fernandez, Rumetshofer, Hochreiter, and Klambauer}]{Sanchez-Fernandez2022.11.17.516915}
Ana Sanchez-Fernandez, Elisabeth Rumetshofer, Sepp Hochreiter, and G{\"u}nter Klambauer. 2023.
\newblock \href {https://doi.org/10.1101/2022.11.17.516915} {CLOOME: contrastive learning unlocks bioimaging databases for queries with chemical structures}.
\newblock \emph{bioRxiv}.

\bibitem[{Schwarzer et~al.(2023)Schwarzer, Obando-Ceron, Courville, Bellemare, Agarwal, and Castro}]{schwarzer2023biggerbetterfasterhumanlevel}
Max Schwarzer, Johan Obando-Ceron, Aaron Courville, Marc Bellemare, Rishabh Agarwal, and Pablo~Samuel Castro. 2023.
\newblock \href {http://arxiv.org/abs/2305.19452} {Bigger, Better, Faster: Human-level Atari with human-level efficiency}.

\bibitem[{Su et~al.(2022)Su, Lan, Wang, Yogatama, Kong, and Collier}]{su2022contrastive}
Yixuan Su, Tian Lan, Yan Wang, Dani Yogatama, Lingpeng Kong, and Nigel Collier. 2022.
\newblock \href {http://arxiv.org/abs/2202.06417} {A Contrastive Framework for Neural Text Generation}.

\bibitem[{Touvron et~al.(2023)Touvron, Lavril, Izacard, Martinet, Lachaux, Lacroix, Rozière, Goyal, Hambro, Azhar, Rodriguez, Joulin, Grave, and Lample}]{touvron2023llama}
Hugo Touvron, Thibaut Lavril, Gautier Izacard, Xavier Martinet, Marie-Anne Lachaux, Timothée Lacroix, Baptiste Rozière, Naman Goyal, Eric Hambro, Faisal Azhar, Aurelien Rodriguez, Armand Joulin, Edouard Grave, and Guillaume Lample. 2023.
\newblock \href {http://arxiv.org/abs/2302.13971} {LLaMA: Open and Efficient Foundation Language Models}.

\bibitem[{van~den Oord et~al.(2019)van~den Oord, Li, and Vinyals}]{oord2019representation}
Aaron van~den Oord, Yazhe Li, and Oriol Vinyals. 2019.
\newblock \href {http://arxiv.org/abs/1807.03748} {Representation Learning with Contrastive Predictive Coding}.

\bibitem[{Wang et~al.(2022{\natexlab{a}})Wang, Wang, Cao, and Barati~Farimani}]{Wang_2022}
Yuyang Wang, Jianren Wang, Zhonglin Cao, and Amir Barati~Farimani. 2022{\natexlab{a}}.
\newblock \href {https://doi.org/10.1038/s42256-022-00447-x} {Molecular contrastive learning of representations via graph neural networks}.
\newblock \emph{Nature Machine Intelligence}, 4(3):279–287.

\bibitem[{Wang et~al.(2022{\natexlab{b}})Wang, Wu, Agarwal, and Sun}]{wang2022medclip}
Zifeng Wang, Zhenbang Wu, Dinesh Agarwal, and Jimeng Sun. 2022{\natexlab{b}}.
\newblock \href {http://arxiv.org/abs/2210.10163} {MedCLIP: Contrastive Learning from Unpaired Medical Images and Text}.

\bibitem[{Wightman et~al.(2021)Wightman, Touvron, and Jégou}]{wightman2021resnet}
Ross Wightman, Hugo Touvron, and Hervé Jégou. 2021.
\newblock \href {http://arxiv.org/abs/2110.00476} {ResNet strikes back: An improved training procedure in timm}.

\bibitem[{Yeh et~al.(2022)Yeh, Hong, Hsu, Liu, Chen, and LeCun}]{yeh2022decoupled}
Chun-Hsiao Yeh, Cheng-Yao Hong, Yen-Chi Hsu, Tyng-Luh Liu, Yubei Chen, and Yann LeCun. 2022.
\newblock \href {http://arxiv.org/abs/2110.06848} {Decoupled Contrastive Learning}.

\bibitem[{You et~al.(2017)You, Gitman, and Ginsburg}]{you2017large}
Yang You, Igor Gitman, and Boris Ginsburg. 2017.
\newblock \href {http://arxiv.org/abs/1708.03888} {Large Batch Training of Convolutional Networks}.

\bibitem[{You et~al.(2021)You, Chen, Sui, Chen, Wang, and Shen}]{you2021graph}
Yuning You, Tianlong Chen, Yongduo Sui, Ting Chen, Zhangyang Wang, and Yang Shen. 2021.
\newblock \href {http://arxiv.org/abs/2010.13902} {Graph Contrastive Learning with Augmentations}.

\bibitem[{Zbontar et~al.(2021)Zbontar, Jing, Misra, LeCun, and Deny}]{zbontar2021barlow}
Jure Zbontar, Li~Jing, Ishan Misra, Yann LeCun, and Stéphane Deny. 2021.
\newblock \href {http://arxiv.org/abs/2103.03230} {Barlow Twins: Self-Supervised Learning via Redundancy Reduction}.

\bibitem[{Zhai et~al.(2023)Zhai, Mustafa, Kolesnikov, and Beyer}]{zhai2023sigmoid}
Xiaohua Zhai, Basil Mustafa, Alexander Kolesnikov, and Lucas Beyer. 2023.
\newblock \href {http://arxiv.org/abs/2303.15343} {Sigmoid Loss for Language Image Pre-Training}.

\bibitem[{Zhou et~al.(2022)Zhou, Dong, Gan, Wang, and Wei}]{zhou2022noncontrastivelearningmeetslanguageimage}
Jinghao Zhou, Li~Dong, Zhe Gan, Lijuan Wang, and Furu Wei. 2022.
\newblock \href {http://arxiv.org/abs/2210.09304} {Non-Contrastive Learning Meets Language-Image Pre-Training}.

\bibitem[{Ömer Veysel~Çağatan(2024)}]{çağatan2024unseeunsupervisednoncontrastivesentence}
Ömer Veysel~Çağatan. 2024.
\newblock \href {http://arxiv.org/abs/2401.15316} {UNSEE: Unsupervised Non-contrastive Sentence Embeddings}.

\end{thebibliography}
\bibliographystyle{Styles/acl_natbib}


\appendix

\section{Appendix}

\subsection{Related Work}\label{sect:related}

\textbf{Sigmoid contrastive learning}

The sigmoid is rather unpopular compared to InfoNCE~\cite{oord2019representation} which greatly promotes the softmax-based contrastive learning however sigmoid has been used in unsupervised dimensionality reduction~\cite{Hadsell2006DimensionalityRB}, supervised learning~\cite{wightman2021resnet,beyer2020imagenet}.

\textbf{SigLIP}

While conventional methodologies rely on softmax normalization, demanding a comparison of each image against the entire dataset, SigLIP~\cite{zhai2023sigmoid} takes a different approach by utilizing a sigmoid loss with learnable temperature and bias. This pairwise loss function directly operates on individual image-text pairs, eliminating the need for a comprehensive assessment of all pairwise similarities. This approach translates to significant enhancements in memory efficiency, allowing for the utilization of larger batch sizes and faster training, even when confronted with limited hardware resources. Notably, SigLIP simplifies implementation compared to softmax-based methods. Despite its straightforward design, models like SigLiT, built upon the foundation of SigLIP, have demonstrated state-of-the-art performance in tasks such as ImageNet zero-shot image classification, outperforming models trained with traditional contrastive learning. The combined benefits of accuracy, efficiency, and simplicity position SigLIP as a compelling alternative for pre-training models that adeptly handle visual-textual relationships

\textbf{Contrastive image pretraining}

SimCLR~\cite{chen2020simple} stands out due to its simplicity and adaptability. It employs a single encoder to extract representations from two augmented versions of an image. The subsequent comparison of these representations involves a temperature-scaled normalized cross-entropy loss. This process encourages the identification of features distinguishing positive pairs from negative ones with the temperature parameter adjusting the difficulty and fostering fine-grained feature discrimination.

On the other hand, MoCo~\cite{he2020momentum} takes a different approach by using a two-encoder system: a query encoder for generating representations and a momentum encoder that gradually catches up to the query. Positive pairs are formed during training as each image passes through both encoders. MoCo uniquely leverages a queue of past representations, enabling the query encoder to learn from a diverse set of examples beyond the current batch. This queue acts as a historical memory bank, enhancing the overall training process.

NNCLR~\cite{dwibedi2021little} acts as a bridge between SimCLR and MoCo, utilizing a single encoder like SimCLR but with a twist in forming positive pairs. Instead of relying on two augmented views of the same image, NNCLR incorporates a memory bank similar to MoCo's queue. However, it identifies positive pairs by locating the k nearest neighbors of the current query representation in the memory bank. This neighbor-based approach focuses on representations sharing similar features, further refining the learned features through a dedicated nearest-neighbor contrastive loss.

\textbf{Non-Contrastive image pretraining}
Recent progress in visual self-supervised learning extends beyond the conventional contrastive paradigm, exploring innovative approaches to diminish reliance on negative samples. These methods are purposefully crafted to enhance the quality of augmented representations, operating independently of negative samples and forming a subset known as non-contrastive objectives. To tackle challenges like model collapse, various strategies have been introduced, including the incorporation of asymmetric architectures~\cite{grill2020bootstrap,chen2020exploring}, utilization of feature decorrelation techniques~\cite{bardes2022vicreg,zbontar2021barlow,ermolov2021whitening,ozsoy2022selfsupervised}, and the integration of clustering methods~\cite{caron2021unsupervised,caron2019deep}.

\subsection{Efficient Chunked Sigmoid Implementation}
The anticipated strength of the sigmoid loss function lies predominantly in distributed training scenarios. While in single-device training, the softmax loss still necessitates two passes of data, it appears to have minimal impact. In contrast, the softmax loss tends to exhibit slightly faster performance, attributed to the inherent masking operations of the sigmoid loss. Although the observed difference is marginal, it becomes more apparent and advantageous when scaling up to distributed training environments. The distinctive characteristics of the sigmoid loss make it particularly well-suited for handling distributed training workloads, where efficiency and scalability are paramount considerations. Thus, we also utilize the efficient chunked sigmoid implementation from SigLIP~\cite{zhai2023sigmoid}.

Contrastive training often utilizes data parallelism, which involves dividing data across $D$ devices. However, computing the loss in this setup requires collecting all embeddings~\cite{zhai2023sigmoid}, which entails costly all-gathers. Moreover, it mandates the creation of a memory-intensive $|\mathcal{B}|\times|\mathcal{B}|$ matrix for pairwise similarities. In contrast, the sigmoid loss is better suited for a memory-efficient, fast, and numerically stable implementation, effectively addressing both challenges. By expressing the per-device batch size as $b = \frac{|\mathcal{B}|}{D}$, the loss can be redefined as:

$$
-\frac 1 {|\mathcal{B}|} 
\underbrace{\sum_{d_i=1}^D}_{\textbf{A: } \forall \text{ device } d_i}
\overbrace{\sum_{d_j=1}^D}^{\substack{\text{\textbf{B:} swap negs} \\ \text{across devices}}}
\overbrace{
\underbrace{\sum_{i=b d_i}^{b(d_i + 1)}}_{\substack{\text{all local} \\ \text{positives}}}
\underbrace{\sum_{j=bd_j}^{b(d_j + 1)}}_{\substack{\text{negs from} \\ \text{next device}}} \mathcal{L}_{ij}}^{\substack{\text{\textbf{C:} per device} \\ \text{loss}}}
$$

This is particularly straightforward for the sigmoid loss since each pair serves as an independent term in the loss function. In essence, we start by calculating the loss component corresponding to the positive pairs and subsequently address $b - 1$ negative pairs in the computation.

Afterward, we permute representations among devices, enabling each device to incorporate negatives from its adjacent device in the subsequent iteration of the summation (\textbf{B}). The loss is subsequently computed to this chunk (sum \textbf{C}). This process is carried out independently on each device, ensuring that each device calculates the loss based on its local batch $b$.

Losses can be efficiently computed by aggregating them across all devices (designated as $\textbf{A}$). The individual collective permutes for the sum ($\textbf{B}$) exhibit fast performance, often surpassing the speed of two all-gathers between $D$ devices. Simultaneously, the memory requirement at any given moment is diminished from $|\mathcal{B}|^2$ to $b^2$ (for the sum $\textbf{C}$). Typically, $b$ remains constant, as the scaling of $|\mathcal{B}|$ is accomplished by increasing the number of accelerators. However, the vanilla loss computation faces rapid scaling bottlenecks due to its quadratic dependency on the batch size~\cite{zhai2023sigmoid}.
\newpage
\subsection{Pseudocode}

\begin{algorithm}
\label{alg:sigclr}
\caption{ Jax-like sigmoid loss pseudocode.}
\newcommand{\hlbox}[1]{%
  \fboxsep=1.2pt\hspace*{-\fboxsep}\colorbox{black!10}{\detokenize{#1}}%
}
\lstset{style=simple}
\begin{lstlisting}[language=Python]
# img_emb: embedding of augmented views [2n, dim] 
# t, b     : fixed temperature and learnable bias
# n        : mini-batch size
# cos_sims : cos_sim of all pairs [2n,2n]
# sim_mask : positive-negative pairs mask
# loss_mask: mask for loss of the same view

diag_range = arange(2n)
shift_range = roll(diag_range,n)

sim_mask = -ones(2n,2n)
loss_mask = ones(2n,2n)
sim_mask[diag_range,shift_range] = 1
loss_mask[diag_range,diag_range] = 0

emb_1, emb2 = img_emb[:,None,:],img_emb[None,:,:]
cos_sims = t * cos_sim(emb_1,emb2) + b

log_lik = -log_sigmoid(cos_sims * sim_mask)
loss = mean(log_lik * loss_mask)


\end{lstlisting}
\end{algorithm}


\end{document}